\def\year{2015}
\documentclass[letterpaper]{article}
\usepackage{aaai}
\usepackage{times}
\usepackage{helvet}
\usepackage{courier}
\usepackage{url}
\usepackage{breakurl}

\usepackage{color}
\usepackage{float}
\usepackage{tabularx}
\usepackage{subfig}
\usepackage{multirow}
\usepackage{dcolumn}
\usepackage{latexsym}
\usepackage{amsmath}
\usepackage{amssymb}
\usepackage{multirow}
\usepackage[lined,boxed,commentsnumbered]{algorithm2e}
\usepackage{xcolor}

\usepackage{mathrsfs}
\usepackage{graphicx}
\usepackage{subfig}
\usepackage{amsfonts}

\usepackage{bm}

\usepackage{tikz-dependency}
\usepackage{pgfplots}
\usetikzlibrary{positioning,bayesnet}
\allowdisplaybreaks

\pgfplotsset{compat=1.5}

\DeclareMathOperator*{\argmax}{arg\,max}

\newcommand\newcite[1]{\citeauthor{#1} (\citeyear{#1})}

\def\bv{\mathbf{v}}

\def\e{\mathbf{e}}

\def\bW{\mathbf{W}}
\def\bC{\mathbf{C}}
\def\bD{\mathbf{D}}

\def\bR{\mathbb{R}}

\def\dN{\mathcal{N}}
\def\dM{\mathcal{M}}

\newenvironment{itemize*}%
  {\begin{itemize}%
    \setlength{\itemsep}{0pt}%
    \setlength{\parskip}{0pt}}%
  {\end{itemize}}
  \newenvironment{enumerate*}%
  {\begin{enumerate}%
    \setlength{\itemsep}{0pt}%
    \setlength{\parskip}{0pt}}%
  {\end{enumerate}}

\begin{document}
%
\title{Gaussian Mixture Embeddings for Multiple Word Prototypes}
\author{Xinchi Chen, Xipeng Qiu\thanks{{ }{ }Corresponding author.}, Jingxiang Jiang, Xuanjing Huang\\
 Shanghai Key Laboratory of Intelligent Information Processing, Fudan University\\
School of Computer Science, Fudan University\\
825 Zhangheng Road, Shanghai, China\\
\{xinchichen13,xpqiu,jxjiang14,xjhuang\}@fudan.edu.cn}
\maketitle
\begin{abstract}
\begin{quote}

Recently, word representation has been increasingly focused on for its excellent properties in representing the word semantics. Previous works mainly suffer from the problem of polysemy phenomenon. To address this problem, most of previous models represent words as multiple distributed vectors. However, it cannot reflect the rich relations between words by representing words as points in the embedded space. In this paper, we propose the Gaussian mixture skip-gram (GMSG) model to learn the Gaussian mixture embeddings for words based on skip-gram framework. Each word can be regarded as a gaussian mixture distribution in the embedded space, and each gaussian component represents a word sense. Since the number of senses varies from word to word, we further propose the Dynamic GMSG (D-GMSG) model by adaptively increasing the sense number of words during training.
Experiments on four benchmarks show the effectiveness of our proposed model.
\end{quote}
\end{abstract}

\section{Introduction}
Distributed word representation has been studied for a considerable efforts \cite{bengio2003neural,morin2005hierarchical,bengio2006neural,Mnih:2007,mikolov2010recurrent,turian2010word,Reisinger:2010,Huang:2012,mikolov2013efficient,mikolov2013distributed}. By representing a word in the embedded space, it could address the problem of curse of dimensionality and capture syntactic and semantic properties. In the embedded space, words that have similar syntactic and semantic roles are also close with each other. Thus, distributed word representation is applied to a abundant natural language processing (NLP) tasks \cite{collobert2008unified,collobert2011natural,socher2011parsing,socher2013parsing}.

Most of previous models map a word as a single point vector in the embedded space, which surfer from polysemy problems. Concretely, many words have different senses in different contextual surroundings. For instance, word ``apple'' could mean a kind of fruit when the contextual information implies that the topic of the text is about food. At the meanwhile, the word ``apple'' could represent the Apple Inc. when the context is about information technology. To address this problem, previous works have gained great success in representing multiple word prototypes. 
\newcite{Reisinger:2010} constructs multiple high-dimensional vectors for each word. \newcite{Huang:2012} learns multiple dense embeddings for each word using global document context.
\newcite{neelakantan2014efficient} proposed multi-sense skip-gram (MSSG) to learn word sense embeddings with online word sense discrimination.
Most of previous works are trying to use multiple points to represent the multiple senses of words, which lead to a drawback. It cannot reflect the rich relations between words by simply representing words as points in the embedded space. For instance, we cannot infer that the word ``fruit'' is the hypernym of the word ``apple'' by representing these two words as two point vectors in the embedded space.




\begin{figure}
\centering
\includegraphics[width=0.45\textwidth]{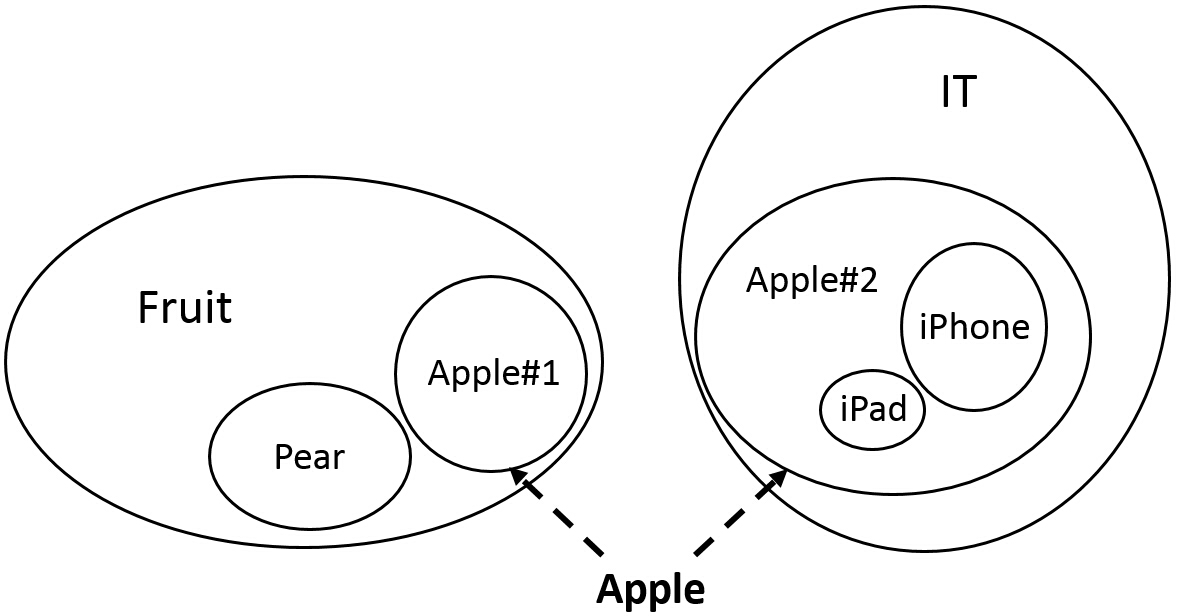}
\caption{An example of Gaussian mixture embeddings.}\label{fig:example_GMSG}
\end{figure}

In this paper, we propose the Gaussian mixture skip-gram (GMSG) model 
inspired by \newcite{vilnis2014word}, who maps a word as a gaussian distribution instead of a point vector in the embedded space.
GMSG model represents each word as a gaussian mixture distribution in the embedded space. Each gaussian component can be regarded as a sense of a word.
Figure \ref{fig:example_GMSG} gives an illustration of Gaussian mixture embeddings.
In this way, much richer relations can be reflected via the relation of two gaussian distributions. For instance, if the word ``fruit'' has larger variance than the word ``apple'', it could show that the word ``fruit'' is the hypernym of the word ``apple''. In addition, using different distance metrics, the relations between words vary. The pair (apple, pear) might be much closer when we use Euclidean distance, while the pair (apple, fruit) might be much closer when we measure them using KL divergence.
Further, since the number of senses varies from word to word, we propose the Dynamic GMSG (D-GMSG) model to handle the varying sense number. D-GMSG automatically increases the sense numbers of words during training. Experiments on four benchmarks show the effectiveness of our proposed models.

\begin{figure*}[t]
  \centering
  \subfloat[Skip-Gram]{
  \includegraphics[width=0.45\linewidth]{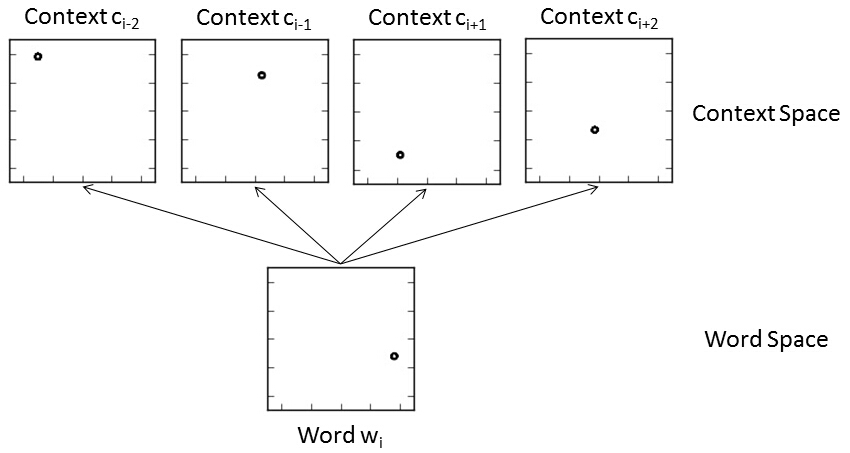} \label{fig:SG}
  }
  \hspace{1em}
  \subfloat[Gaussian Skip-Gram]{
  \includegraphics[width=0.45\linewidth]{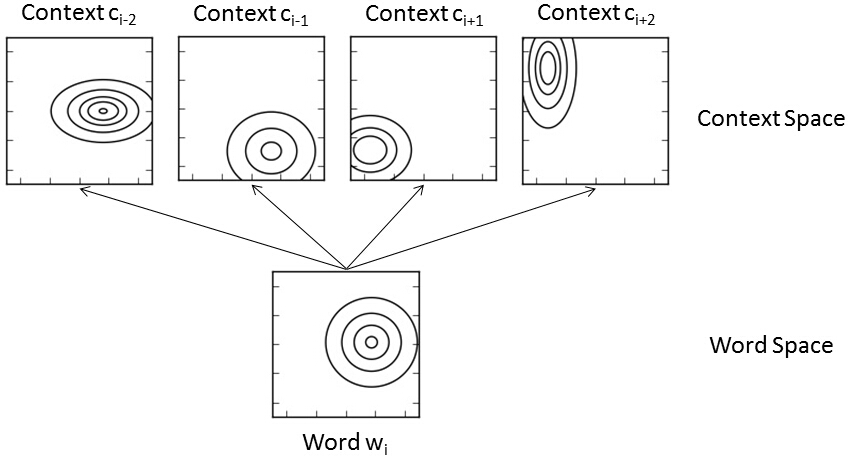} \label{fig:GSG}
  }\\
  \subfloat[Gaussian Mixture Skip-Gram]{
  \includegraphics[width=0.45\linewidth]{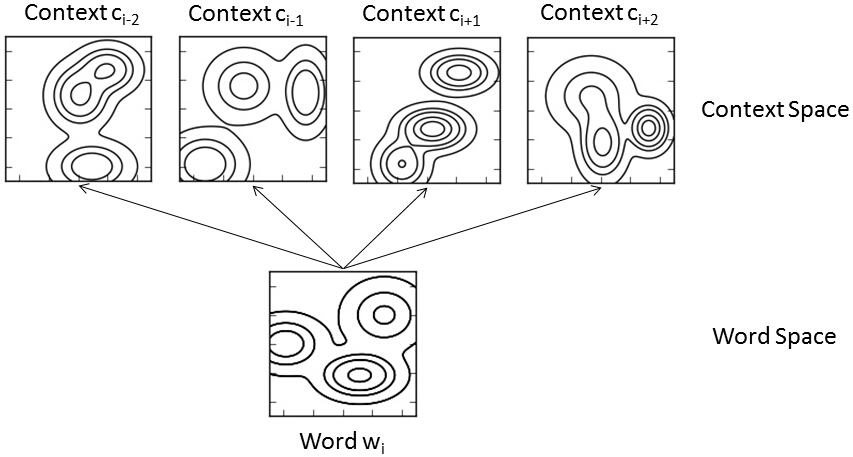} \label{fig:GMSG}
  }
  \hspace{1em}
  \subfloat[Dynamic Gaussian Mixture Skip-Gram]{
  \includegraphics[width=0.45\linewidth]{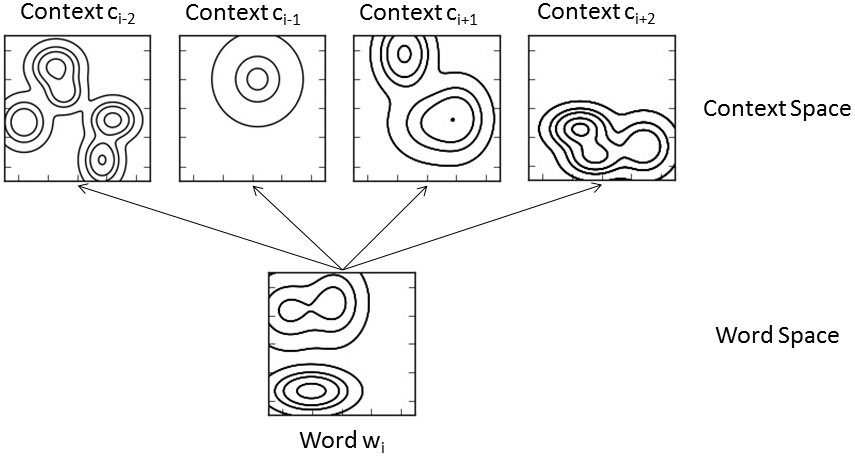} \label{fig:D-GMSG}
  }
  \caption{Four architectures for learning word representations.}\label{fig:ours}
\end{figure*}
\section{Skip-Gram}
In the skip-gram
model (Figure \ref{fig:SG}) \cite{Mikolov:2013}, each word $w$ is represented as a distributed vector $\e_w \in \bW$, where $\bW \in \bR^{|V| * d}$ is the word embedding matrix for all words in word vocabulary $V$. $d$ is the dimensionality of the word embeddings. Correspondingly, each context word $c_w$ also has a distributed representation vector $\hat{\e}_{c_w} \in \bC$, where $\bC \in \bR^{|V| * d}$ is another distinguished space.

The skip-gram aims to maximize the probability of the co-occurrence of a word $w$ and its context word $c_w$, which can be formalized as:
\begin{equation}
  \bv(w|c_w) = \prod_{u \in \{w\} \bigcup NEG(w)} p(u|c_w),
\end{equation}
where function $NEG(\cdot)$ returns a set of negative sampling context words and $p(u|c_w)$ can be formalized as:
\begin{equation}
p(u|c_w) = [\sigma(\e_w^\top \hat{\e}_u)]^{\textbf{1}\{u=c_w\}} [1 - \sigma(\e_w^\top \hat{\e}_u)]^{\textbf{1}\{u \neq c_w\}},
\end{equation}
where $\textbf{1}\{\cdot\}$ is a indicator function and $\sigma(\cdot)$ is the sigmoid function.

Given a training corpus $\bD$ which insists of $(word, context)$ pairs $(w, c_w) \in \bD$, the goal of skip-gram model is to maximize the objective function:
\begin{equation}
  J(\theta) = \sum_{(w, c_w) \in \bD} \log \bv(w|c_w),
\end{equation}
where $\theta$ is the parameter set of the model.

Concretely, the $NEG(w)$ function samples negative contextual words for current word $w$ drawn on the distribution:
\begin{equation}
  P(w) \propto P_{unigram}(w)^{sr},
\end{equation}
where $P_{unigram}(w)$ is a unigram distribution of words and $sr$ is a hyper-parameter.

%
%

\section{Gaussian Skip-Gram Model}
Although the skip-gram model is extremely efficient and the learned word embeddings have greet properties on the syntactic and semantic roles, it cannot give the asymmetric distance between words.

Skip-gram model defines the representation of a word as a vector, and defines the similarity of two words $w_1$ and $w_2$ using cosine distance of vectors $\e_{w_{1}}$ and $\e_{w_{2}}$.
Unlike the definition of skip-gram model, in this paper, we argue that a word can be regarded as a function. The similarity between two function $f$ and $g$ can be formalized as:
\begin{equation}
  sim(f,g) = \int_{x \in \bR^n} f(x) g(x) dx
\end{equation}

Specifically, when we choose the gaussian distribution as the function,
\begin{gather}\small
f(x) = \dN ( x ; \mu_\alpha, \Sigma_\alpha),\\
g(x) = \dN ( x ; \mu_\beta, \Sigma_\beta),
\end{gather}
the similarity between two gaussian distributions $f$ and $g$ can be formalized as:
\begin{gather}\small
  sim(f,g) =   \int_{x \in \bR^n} \dN ( x ; \mu_\alpha, \Sigma_\alpha) \dN ( x ; \mu_\beta, \Sigma_\beta) dx \\
  = \dN ( 0; \mu_\alpha - \mu_\alpha, \Sigma_\beta + \Sigma_\beta),
\end{gather}
where $\mu_\alpha$, $\mu_\beta$ and $\Sigma_\alpha$, $\Sigma_\beta$ are the mean and covariance matrix of word $f$ and word $g$ respectively.

In this way, we can use KL divergence to measure the distance of two words which are represented as distributions. Since KL divergence is not symmetric metric, Gaussian skip-gram (GSG) model (Figure \ref{fig:GSG}) \cite{vilnis2014word} could measure the distance between words asymmetrically.
\section{Gaussian Mixture Skip-Gram Model}
Although GSG model seems work well, it cannot handle the problem of polysemy phenomenon of words. In this paper, we propose the Gaussian mixture skip-gram (GMSG) model (Figure \ref{fig:GMSG}). GMSG regard each word as a gaussian mixture distribution. Each component of the gaussian mixture distribution of a word can be regarded as a sense of the word.
The senses are automatically learned during training using the information of occurrences of words and their context. Besides handling the polysemy problem, GMSG also captures the richer relations between words. As shown in Figure \ref{fig:example_GMSG}, it is tricky to tell whose distance is smaller between word pairs (apple, fruit) and (apple, pear). On the one hand, word ``apple'' seems much closer with word ``fruit'', in a manner, since apple is a kind of fruit while ``pear'' and ``apple'' are just syntagmatic relation. On the other hand, word ``apple'' seems much closer with word ``pear'', since they are in the same level of semantic granularity. Actually, in different distance metrics, the relations between words varies. The pair (apple, pear) might be much closer when we use Euclidean distance, while the pair (apple, fruit) might be much closer when we measure them using KL divergence.
However, whatever the relationship between them are, their representations are learned and fixed.

Formally, we define the distributions of two words as $f$ and $g$:
\begin{gather}\small
f(x) = \dN ( x ; \phi_\alpha, \mu_\alpha, \Sigma_\alpha), \\
g(x) = \dN ( x ; \phi_\beta, \mu_\beta, \Sigma_\beta),
\end{gather}
where $\phi_\alpha$ and $\phi_\beta$ are the parameters of multinomial distributions.

The similarity of two distributions can be formalized as:
\begin{gather}\small
  sim(f,g) =   \int_{x \in \bR^n} \dN ( x ; \phi_\alpha, \mu_\alpha, \Sigma_\alpha) \dN ( x ; \phi_\beta, \mu_\beta, \Sigma_\beta) dx \\
=  \int_{x \in \bR^n} \left[ \sum_{i}\dN ( x | z = i; \mu_\alpha, \Sigma_\alpha) \dM (z = i; \phi_\alpha) \right] \times \\
\left[ \sum_{j} \dN ( x | z = j; \mu_\beta, \Sigma_\beta) \dM (z = i; \phi_\beta)  \right] dx\\
=\int_{x \in \bR^n} \left[ \sum_{i}\dN ( x; \mu_{\alpha i}, \Sigma_{\alpha i}) \phi_{\alpha i}) \right] \times \\
\left[ \sum_{j} \dN ( x; \mu_{\beta j}, \Sigma_{\beta j}) \phi_{\beta j})  \right] dx\\
= \sum_{i} \sum_{j} \phi_{\alpha i} \phi_{\beta j} \int_{x \in \bR^n} \dN ( x; \mu_{\alpha i}, \Sigma_{\alpha i}) \dN ( x; \mu_{\beta j}, \Sigma_{\beta j}) dx \\
  = \sum_{i} \sum_{j} \phi_{\alpha i} \phi_{\beta j} \dN ( 0; \mu_{\alpha i} - \mu_{\beta j}, \Sigma_{\alpha i} + \Sigma_{\beta j}),
\end{gather}
where $\dM$ represents multinomial distribution.

Algorithm \ref{alg:GMSG} shows the details, where $\phi$, $\mu$ and $\Sigma$ are parameters of word representations in word space. $\hat{\phi}$, $\hat{\mu}$ and $\hat{\Sigma}$ are parameters of word representations in context space.
\begin{algorithm}[ht!]

\SetKwInOut{Input}{Input}
\SetKwInOut{Output}{Output}
\Input{Training corpus: $w_1$, $w_2$, $\dots$, $w_T$;\\
Sense number: $K$;\\
Dimensionality of mean: $d$;\\
Context window size: $N$;\\
Max-margin: $\kappa$.}

\BlankLine
\textbf{Initialize}: Multinomial parameter: $\phi, \hat{\phi} \in \bR^{|V| \times K}$;\\
Means of Gaussian mixture distribution: $\mu, \hat{\mu} \in \bR^{|V| \times d}$;\\
Covariances of Gaussian mixture distribution: $\Sigma, \hat{\Sigma} \in \bR^{|V| \times d \times d}$;\\

\For{$w = w_1 \cdots w_T$}{
    $n_w$ $\sim$ \{1, $\dots$, $N$\}\\
    $c(w)$ = \{$w_{t-n_w}$, $\dots$, $w_{t-1}$, $w_{t+1}$, $\dots$, $w_{t+n_w}$\}\\
    \For{$c$ in $c(w)$}{
        \For{$\hat{c}$ in $NEG(\hat{c})$}{
            $l = \kappa - sim(f_w, f_c) + sim(f_w, f_{\hat{c}})$\;
            \If {$l > 0$} {
                Accumulate gradient for $\phi_w$, $\mu_w$, $\Sigma_w$\;
                Gradient update on $\hat{\phi}_c$, $\hat{\phi}_{\hat{c}}$,
                $\hat{\mu}_c$, $\hat{\mu}_{\hat{c}}$,
                $\hat{\Sigma}_c$, $\hat{\Sigma}_{\hat{c}}$\;
            }
            Gradient update for $L_2$ normalization term of $\hat{\phi}_c$, $\hat{\phi}_{\hat{c}}$,
                $\hat{\mu}_c$, $\hat{\mu}_{\hat{c}}$,
                $\hat{\Sigma}_c$, $\hat{\Sigma}_{\hat{c}}$\;
        }
        Gradient update on $\phi_w$, $\mu_w$, $\Sigma_w$\;
        Gradient update for $L_2$ normalization term of $\phi_w$, $\mu_w$, $\Sigma_w$\;
    }
}
\BlankLine
\Output{$\phi$, $\mu$ and $\Sigma$}
\caption{Training algorithm of GMSG model using max-margin criterion.}\label{alg:GMSG}
\end{algorithm}

\section{Dynamic Gaussian Mixture Skip-Gram Model}
Although we could represent polysemy of words by using gaussian mixture distribution, there is still a short that should be pointed out. Actually, the number of word senses varies from word to word. To dynamically increasing the numbers of gaussian components of words, we propose the Dynamic Gaussian Mixture Skip-Gram (D-GMSG) model (Figure \ref{fig:D-GMSG}). The number of senses for a word is unknown and is automatically learned during training.

At the beginning, each word is assigned a random gaussian distribution. During training, a new gaussian component will be generated when the similarity of the word and its context is less than $\gamma$, where $\gamma$ is a hyper-parameter of D-GMSG model.

Concretely, consider the word $w$ and its context $c(w)$. The similarity of them is defined as:
\begin{equation}
  s(w,c(w)) = \frac{1}{|c(w)|}\sum_{c \in c(w)} sim(f_w,f_c).
\end{equation}

For word $w$, assuming that $w$ already has $k$ Gaussian components, it is represented as
\begin{gather}
\begin{aligned}
f_w &= \dN(\cdot;\phi_w,\mu_w,\Sigma_w),\\
  \phi_w &= \{\phi_{w_i}\}_{i=1}^{k},\\
  \mu_w &= \{\mu_{w_i}\}_{i=1}^{k},\\
  \Sigma_w &= \{\Sigma_{w_i}\}_{i=1}^{k}.
\end{aligned}
\end{gather}

When $s(w,c(w)) < \gamma$, we generate a new random gaussian component $\dN(\cdot;\mu_{w_{k+1}},\Sigma_{w_{k+1}})$ for word $w$, and $f_w$ is then updated as
\begin{equation}
\begin{aligned}
f_w^* &= \dN(\cdot;\phi_w^*,\mu_w^*,\Sigma_w^*), \\
\phi_w^* &= \{(1 - \xi)\phi_{w_i}\}_{i=1}^{k} \oplus \xi, \\
\mu_w^* &= \{\mu_{w_i}\}_{i=1}^{k} \oplus \mu_{w_{k+1}}, \\
\Sigma_w^* &= \{\Sigma_{w_i}\}_{i=1}^{k} \oplus \Sigma_{w_{k+1}},
\end{aligned}
\end{equation}
where mixture coefficient $\xi$ is a hyper-parameter and operator $\oplus$ is an set union operation.

\section{Relation with Skip-Gram}
In this section, we would like to introduce the relation between GMSG model and prevalent skip-gram model.

Skip-gram model \cite{Mikolov:2013} is a well known model for learning word embeddings for its efficiency and effectiveness. Skip-gram defines the similarity of word $w$ and its context $c$ as: 
\begin{equation}
  \begin{aligned}
    sim(w,c) 
            &= \frac{1}{1+\exp(-\e_w^\top \hat{\e}_c)}\\
  \end{aligned}
\end{equation}

When the covariances of all words and context words $\Sigma_w = \hat{\Sigma}_c = \frac{1}{2} \mathbf{I}$ are fixed, $\mu_{f_w} = \e_w$, $\mu_{f_c} = \hat{\e}_c$ and sense number $K = 1$, the similarity of GMSG model can be formalized as:
\begin{equation}
\begin{aligned}
  sim(f_w,f_c) &= \dN(0; \mu_{f_w} - \mu_{f_c}, \Sigma_{f_w} + \Sigma_{f_c} ) \\
  &= \dN(0; \e_w - \hat{\e}_c, \Sigma_w + \hat{\Sigma}_c ) \\
  &= \dN(0; \e_w - \hat{\e}_c, \mathbf{I})\\
  &\propto \exp ((\e_w - \hat{\e}_c)^\top (\e_w - \hat{\e}_c)).
  \end{aligned}
\end{equation}

The definition of similarity of skip-gram is a function of dot product of $\e_w$ and $\hat{\e}_c$, while it is a function of Euclidian distance of $\e_w$ and $\hat{\e}_c$. In a manner, skip-gram model is a related model of GMSG model conditioned on fixed $\Sigma_w = \hat{\Sigma}_c = \frac{1}{2} \mathbf{I}$, $\mu_{f_w} = \e_w$, $\mu_{f_c} = \hat{\e}_c$ and $K = 1$.

\section{Training}
The similarity function of the skip-gram model is dot product of vectors, which could perform a binary classifier to tell the positive and negative $(word, context)$ pairs.
In this paper, we use a more complex similarity function, which defines a absolute value for each positive and negative pair rather than a relative relation. Thus, we minimise a different max-margin based loss function $L(\theta)$ following \cite{Joachims:2002} and \cite{weston2011wsabie}:
\begin{equation}\small
\begin{aligned}
    L(\theta) &= \frac{1}{Z} \sum_{(w,c_w) \in \bD} \sum_{c^- \in NEG(w)} l(\theta) + \frac{1}{2} \lambda \|\theta\|^2_2,\\
    l(\theta) &= \max\{0, \kappa - sim(f_w,f_{c_w}) + sim(f_w, f_{c^-})\},
\end{aligned}\label{eq:loss}
\end{equation}
where $\theta$ is the parameter set of our model and the margin $\kappa$ is a hyper-parameter. $Z = \sum_{(w,c_w) \in \bD} |NEG(w)|$ is a normalization term. 
Conventionally, we add a $L_2$ regularization term for all parameters, weighted by a hyper-parameter $\lambda$.

Since the objective function is not differentiable due
to the hinge loss, we use the sub-gradient method \cite{duchi2011adaptive}. Thus, the subgradient of Eq. \ref{eq:loss} is:
\begin{equation}
\begin{aligned}
\frac{\partial L(\theta)}{\partial \theta} &= \frac{1}{Z}\sum_{(w,c_w) \in \bD} \sum_{c^- \in NEG(w)} \frac{\partial l(\theta)}{\partial \theta} + \lambda\theta, \\
  \frac{\partial l(\theta)}{\partial \theta} &= -\frac{\partial sim(f_w,f_{c_w})}{\partial \theta} + \frac{\partial sim(f_w,f_{c^-})}{\partial \theta}.
\end{aligned}
\end{equation}

In addition, the covariance matrices need to be kept positive definite. Following \newcite{vilnis2014word}, we use diagonal covariance matrices with a hard constraint that the eigenvalues $\varrho$ of the covariance matrices lie within the interval $[m,M]$.



\section{Experiments}
To evaluate our proposed methods, we learn the word representation using the Wikipedia corpus\footnote{\url{http://mattmahoney.net/dc/enwik9.zip}}. We experiment on four different benchmarks: WordSim-353, Rel-122, MC and SCWS. Only SCWS provides the contextual information.
\paragraph{WordSim-353}
WordSim-353\footnote{\url{http://www.cs.technion.ac.il/~gabr/resources/data/wordsim353/wordsim353.html}}
\cite{finkelstein2001placing} consists of 353 pairs of words and their similarity scores.
\paragraph{Rel-122}
Rel-122\footnote{\url{http://www.cs.ucf.edu/~seansz/rel-122/}}
\cite{szumlanski2013new} contains 122 pairs of nouns and compiled them into a
new set of relatedness norms. 
\paragraph{MC}
MC\footnote{\url{http://www.cs.cmu.edu/~mfaruqui/word-sim/EN-MC-30.txt}}
\cite{miller1991contextual} contains 30 pairs of nouns that vary from high to low semantic similarity. 
\paragraph{SCWS}
SCWS\footnote{\url{http://www.socher.org/index.php/Main/ImprovingWordRepresentationsViaGlobalContextAndMultipleWordPrototypes}}
\cite{Huang:2012} consists of 2003 word pairs and their contextual information. Concretely, the dataset consists of 1328 noun-noun, 97 adjective-adjective, 30 noun-adjective, 9 verb-adjective, 399 verb-verb, 140 verb-noun and 241 same-word pairs.
\subsection{Hyper-parameters}
The hyper-parameter settings are listed in the Table \ref{tab:hyper-parameter}. In this paper, we evaluate our models conditioned on the dimensionality of means $d = 50$, and we use diagonal covariance matrices for experiments. We remove all the word with occurrences less than 5 (Min count).

\begin{table}[!t] \small 
\centering
\begin{tabular}{|l|l|}
    \hline
    Sampling rate&$sr = 3/4$\\
    Context window size &$N = 5$\\
    Dimensionality of mean&$d = 50$\\
    Initial learning rate&$\alpha = 0.025$\\
    Margin&$\kappa = 0.5$\\
    Regularization&$\lambda = 10^{-8}$\\
    Min count&$mc = 5$\\
    Number of learning iteration&$iter = 5$\\
    Sense Number for GMSG&$K = 3$\\
    Weight of new Gaussian component&$\xi = 0.2$\\
    Threshold for generating a new Gaussian component&$\gamma = 0.02$\\
    \hline
\end{tabular}
\caption{Hyper-parameter settings.}\label{tab:hyper-parameter}
\end{table}
\subsection{Model Selection}
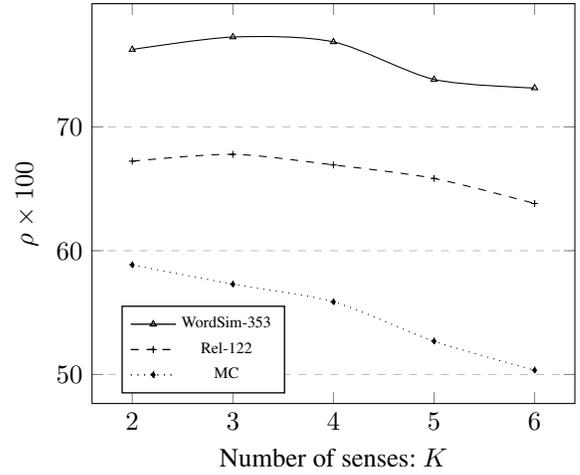
\begin{figure}[t]
  \centering
  \pgfplotsset{width=0.45\textwidth}
  \pgfplotscreateplotcyclelist{my black white}{%
, every mark/.append style={solid, fill=gray}, \\%
dotted, every mark/.append style={solid, fill=gray}, mark=square*\\%
densely dotted, every mark/.append style={solid, fill=gray}, mark=otimes*\\%
loosely dotted, every mark/.append style={solid, fill=gray}, mark=triangle*\\%
dashed, every mark/.append style={solid, fill=gray},mark=diamond*\\%
loosely dashed, every mark/.append style={solid, fill=gray},mark=*\\%
densely dashed, every mark/.append style={solid, fill=gray},mark=square*\\%
dashdotted, every mark/.append style={solid, fill=gray},mark=otimes*\\%
dasdotdotted, every mark/.append style={solid},mark=star\\%
densely dashdotted,every mark/.append style={solid, fill=gray},mark=diamond*\\%
}
  \begin{tikzpicture}
    \begin{axis}[
    xlabel={Number of senses: $K$},
    ylabel={$\rho \times 100$},
    legend entries={WordSim-353, Rel-122, MC},
    mark size=1.3pt,
    ymajorgrids=true,
    grid style=dashed,
    legend style={font=\tiny,line width=.5pt,mark size=1.3pt,
            at={(0.40,0.135)},
            anchor=east,
            /tikz/every even column/.append style={column sep=0.5em}},
            smooth,
    ]
    \addplot [black,solid,mark=triangle] table [x index=0, y index=3] {modelSelect100.txt};
    \addplot [black,dashed,mark=+] table [x index=0, y index=1] {modelSelect100.txt};
    \addplot [black,dotted,mark=diamond*] table [x index=0, y index=2] {modelSelect100.txt};
    \end{axis}
\end{tikzpicture}
  \caption{Performance of GMSG using different sense numbers.}\label{fig:model_select}
\end{figure}

Figure \ref{fig:model_select} shows the performances using different sense number for words. According to the results, sense number $sn = 3$ for GMSG model is a good trade off between efficiency and model performance.

\subsection{Word Similarity and Polysemy Phenomenon}
To evaluate our proposed models, we experiment on four benchmarks, which can be divided into two kinds. Datasets (Sim-353, Rel-122 and MC) only provide the word pairs and their similarity scores, while SCWS dataset additionally provides the contexts of word pairs. It is natural way to tackle the polysemy problem using contextual information, which means a word in different contextual surroundings might have different senses.

Table \ref{tab:res_uni} shows the results on Sim-353, Rel-122 and MC datasets, which shows that our models have excellent performance on word similarity tasks. Table \ref{tab:res_SCWS} shows the results on SCWS dataset, which shows that our models perform well on polysemy phenomenon.

In this paper, we define the similarity of two words $w$ and $w'$ as: 
\begin{equation}\small
\begin{aligned}
    AvgSim(f_w,f_{w'}) &= sim(f_w,f_{w'}) \\
                       &= \sum_i \sum_j \phi_{w i} \phi_{w' j} sim(f_{w i},f_{w' j}),
\end{aligned}
\end{equation}
where $f_{w}$ and $f_{w'}$ are the distribution representations of the corresponding words. Table \ref{tab:res_uni} gives the results on the three benchmarks. The size of word representation of all the previous models are chosen to be 50 in this paper.

\begin{table}[!t] \small
\centering
%
\begin{tabular}{|l|*{3}{c|}}
\hline
\textbf{Methods}&\textbf{WordSim-353}&\textbf{Rel-122}&\textbf{MC}\\
  \hline
  SkipGram&   59.89    &   49.14&63.96\\
  \hline
    MSSG &	 63.2&	 - & -\\
  NP-MSSG&	 62.4&	 - & -\\
  \hline

 GSG-S&62.03&51.09&69.17\\
  GSG-D&61.00&53.54&68.50\\
     \hline
  GMSG&\textbf{67.8}&\textbf{57.3}&\textbf{77.3}\\
  D-GMSG&56.3&47.1&47.3\\
\hline
\end{tabular}
\caption{Performances on WordSim-353, Rel-122 and MC datasets. MSSG \cite{neelakantan2014efficient} model sets the same sense number for each word. NP-MSSG model automatically learns the sense number for each word during training. GSG \cite{vilnis2014word} has several variations. ``S'' indicates the GSG uses spherical covariance matrices. ``D'' indicates the GSG uses diagonal covariance matrices.}\label{tab:res_uni}
\end{table}

To tackle the polysemy problem, we incorporate the contextual information. In this paper, we define the similarity of two words ($w$, $w'$) with their contexts ($c(w)$, $c(w')$) as:
\begin{equation}\small
    MaxSimC(f_w,f_{w'}) = sim(f_{w k},f_{w' k'})
\end{equation}
where $k = \argmax_i P(i|w,c(w))$ and $k' = \argmax_j P(j|w',c(w'))$. Here, $P(i|w,c(w))$ gives the probability of the $i$-th sense of the current word $w$ will take conditioned on the specific contextual surrounding $c(w)$, where $P(i|w,c(w))$ is defined as:
\begin{equation}\small
    P(i|w,c(w)) = \frac{\frac{1}{|c(w)|} \sum_{c \in c(w)} \phi_{w i} sim(f_{w i}, f_{c})}{\sum_j \frac{1}{|c(w)|} \sum_{c' \in c(w)} \phi_{w j} sim(f_{w j}, f_{c'})}.
\end{equation}

Table \ref{tab:res_SCWS} gives the results on the SCWS benchmark.

\begin{figure}
\centering
\includegraphics[width=0.45\textwidth]{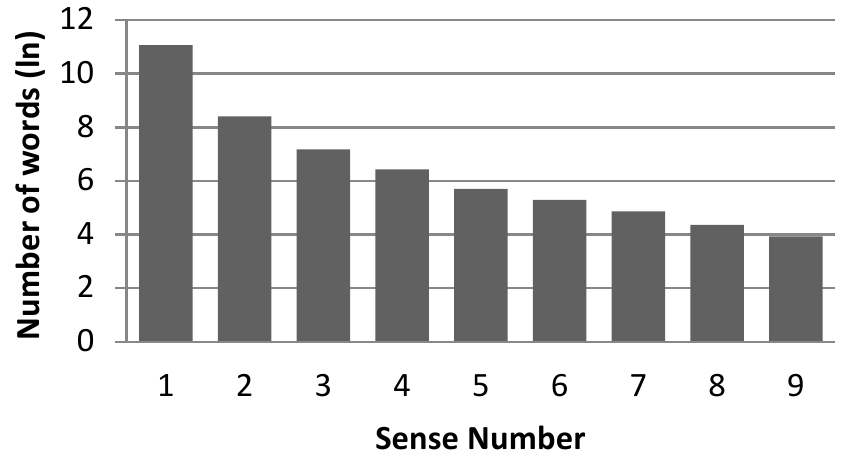}
\caption{Distribution of sense numbers of words learned by D-GMSG.}\label{fig:Num_D-GMSG}
\end{figure}

Figure \ref{fig:Num_D-GMSG} gives the distribution of sense numbers of words using logarithmic scale, which is trained by D-GMSG model. The vocabulary size is $71084$ here. As shown in Figure \ref{fig:Num_D-GMSG}, majority of words have only one sense, and the number of words decreases progressively with the increase of word sense number.

\begin{table}[!t]\small
\centering
\begin{tabular}{|l|c|c|}
\hline
\textbf{Model} & \multicolumn{2}{c|}{$\rho$ $\times$ 100}  \\
\hline
Pruned TFIDF & \multicolumn{2}{c|}{62.5}  \\
Skip-Gram & \multicolumn{2}{c|}{63.4}  \\
C\&W & \multicolumn{2}{c|}{57.0}  \\
\hline
& \textbf{AvgSim}& \textbf{MaxSimC} \\
\hline

  Huang&62.8&26.1\\
  MSSG&64.2&49.17\\
  NP-MSSG&64.0&50.27\\
\hline
\textbf{our models} &&\\
  GMSG&\textbf{64.6}&\textbf{53.6}\\
  D-GMSG&56.0&41.4\\
\hline
\end{tabular}
\caption{Performance on SCWS dataset.Pruned TFIDF \cite{Reisinger:2010} uses spare, high-dimensional word representations. C\&W \cite{collobert2008unified} is a language model. Huang \cite{Huang:2012} is a neural network model for learning multi-representations per word.}\label{tab:res_SCWS}
\end{table}

\section{Related Work}
Recently, it has attracted lots of interests to learn word representation.
Much previous works focus on learning word as a point vector in the embedded space. \newcite{bengio2003neural} applies the neural network to sentence modelling instead of using $n$-gram language models. Since the \newcite{bengio2003neural} model need expensive computational resource. Lots of works are trying to optimize it. \newcite{mikolov2013efficient} proposed the skip-gram model which is extremely efficient by removing the hidden layers of the neural networks, so that larger corpus could be used to train the word representation. By representing a word as a distributed vector, we gain a lot of interesting properties, such as similar words in syntactic or semantic roles are also close with each other in the embedded space in cosine distance or Euclidian distance. In addition, word embeddings also perform excellently in analogy tasks. For instance, $\e_{king} - \e_{queen} \approx \e_{man} - \e_{woman}$.

However, previous models mainly suffer from the polysemy problem. To address this problem, 
\newcite{Reisinger:2010} represents words by constructing multiple sparse, high-dimensional vectors. \newcite{Huang:2012} is an neural network based approach, which learns multiple dense, low-dimensional embeddings using global document context. 
\newcite{Tian:2014} modeled the word polysemy from a probabilistic perspective and integrate it with the highly efficient continuous Skip-Gram model.
\newcite{neelakantan2014efficient} proposed Multi-Sense Skip-Gram (MSSG) to learn word sense embeddings with online word sense discrimination.
These models perform word sense discrimination by clustering context of words. \newcite{liu2015topical} discriminates word sense by introducing latent topic model to globally cluster the words into different topics. \newcite{liulearning} further extended this work to model the complicated interactions of word embedding and its corresponding topic embedding by incorporating the tensor method.

Almost previous works are trying to use multiple points to represent the multiple senses of words. However, it cannot reflect the rich relations between words by simply representing words as points in the embedded space.
\newcite{vilnis2014word} represented a word as a gaussian distribution. 
Gaussian mixture skip-gram (GMSG) model represents a word as a gaussian mixture distribution. Each sense of a word can be regarded as a gaussian component of the word. 
GMSG model gives different relations of words under different distance metrics, such as cosine distance, dot product, Euclidian distance KL divergence, etc.
\section{Conclusions and Further Works}
In this paper, we propose the Gaussian mixture skip-gram (GMSG) model to map a word as a density in the embedded space. A word is represented as a gaussian mixture distribution whose components can be regarded as the senses of the word. GMSG could reflect the rich relations of words when using different distance metrics. 
Since the number of word senses varies from word to word, we further propose the Dynamic GMSG (D-GMSG) model to adaptively increase the sense numbers of words during training.

Actually, a word can be regarded as any function including gaussian mixture distribution. In the further, we would like to investigate the properties of other functions for word representations and try to figure out the nature of the word semantic.
\bibliographystyle{aaai}
\bibliography{./GMSG}

\end{document}